\documentclass{llncs}

\usepackage{color}

\usepackage{url}

\usepackage{listings}
\usepackage{varwidth}

\usepackage{graphicx}

\makeatletter
\newcommand{\@BIBLABEL}{\@emptybiblabel}
\newcommand{\@emptybiblabel}[1]{}
\makeatother

\usepackage{hyperref}

\def\concat#1#2#3{#1#2#3}
\usepackage{enumitem}

\usepackage{helvet}
\usepackage[table]{xcolor}
\usepackage{soul}
\usepackage[titletoc,toc,title]{appendix}

\usepackage{amsmath}

\begin{document}

\title{Providing Self-Aware Systems with Reflexivity}

\author{Alessandro Valitutti\inst{1} \and Giuseppe Trautteur\inst{2}}

\institute{University of Bari
\and
University of Naples Federico II}

\maketitle

%% TMP COMMENT
%% Modify the bibliography environment to call for the author-year
%% system. This is done normally with the citeauthoryear option
%% for a particular contribution.
%\makeatletter
%\renewenvironment{thebibliography}[1]
%     {\section*{\refname}
%      \small
%      \list{}%
%           {\settowidth\labelwidth{}%
%            \leftmargin\parindent
%            \itemindent=-\parindent
%            \labelsep=\z@
%            \if@openbib
%              \advance\leftmargin\bibindent
%              \itemindent -\bibindent
%              \listparindent \itemindent
%              \parsep \z@
%            \fi
%            \usecounter{enumiv}%
%            \let\p@enumiv\@empty
%            \renewcommand\theenumiv{}}%
%      \if@openbib
%        \renewcommand\newblock{\par}%
%      \else
%        \renewcommand\newblock{\hskip .11em \@plus.33em \@minus.07em}%
%      \fi
%      \sloppy\clubpenalty4000\widowpenalty4000%
%      \sfcode`\.=\@m}
%     {\def\@noitemerr
%       {\@latex@warning{Empty `thebibliography' environment}}%
%      \endlist}
%      \def\@cite#1{#1}%
%      \def\@lbibitem[#1]#2{\item[]\if@filesw
%        {\def\protect##1{\string ##1\space}\immediate
%      \write\@auxout{\string\bibcite{#2}{#1}}}\fi\ignorespaces}
%\makeatother

\begin{abstract}

We propose a new type of self-aware systems inspired by ideas from higher-order theories of consciousness. First, we discussed the crucial distinction between introspection and reflexion. Then, we focus on computational reflexion as a mechanism by which a computer program can inspect its own code at every stage of the computation. Finally, we provide a formal definition and a proof-of-concept implementation of computational reflexion, viewed as an enriched form of program interpretation and a way to dynamically ``augment" a computational process.


\end{abstract}

%-------------------------------------------
\section{Introduction}
\label{introduction}

%\parared{Definition of self-aware computing}

Self-aware computing is a recent area of computer science concerning autonomic computing systems capable of capturing knowledge about themselves, maintaining it, and using it to perform self-adaptive behaviors at runtime\cite{Lewis_et_al2015}\cite{Torresen_et_al2015}\cite{Amir_et_al2004}.
%\parared{Three properties}
Almost all self-aware systems share one or more of three properties dealt with extensively in the AI literature: \emph{self-representation}, \emph{self-modification}, and \emph{persistence}.
%\parared{Examples of self-aware behaviors}
Examples of self-aware behaviors are the introspection and reflection features implemented in some programming languages such as Java. Type introspection is the ability of a program to examine the type or properties of an object at runtime, while reflection\footnote{The term \emph{reflection} should not be confused with the term \emph{reflexion}, which will be discussed in Sections \ref{conscious-reflexivity} and \ref{computational-reflexivity}.}  additionally allows a program to manipulate objects and functions at runtime.
 
%\parared{No all properties satisfied}
However, neither of them have all of the above three properties. In fact, introspection implies self-representation but not self-modification. Moreover, reflection is temporally-bound, since it occurs in a small portion of the program execution. Even self-monitoring, considered as a periodic sequence of introspective events, implies persistence but not self-modification.
%\parared{Research question}
We may wonder if we could have a type of computational self-awareness in which persistent self-representation and self-modification would occur simultaneously and yet being functionally distinct. 

%\parared{Paper contribution}

In this paper, we address this issue and present a computational architecture provided with this property, which we call \emph{computational reflexivity}.
%\parared{Specific strategy: enrichment of the execution loop}
Specifically, we propose to introduce introspection and reflection at every step of the execution, enriching the interpretation loop with additional instructions aimed to represent the program at a meta level, combine local and global information, and perform a second-order execution.
%\parared{Concurrent execution}
The enriched interpreter is thus capable of running a program and, concurrently, generating and executing a corresponding modified (or ``augmented'') version.

%\parared{Virtual modification as main advantage}

This separation between ``observed'' (or \emph{target}) and ``observing'' (or \emph{augmented}) process allows the system to perform self-modification at a virtual level (i.e., on the augmented process). As a consequence, the system can choose whether and when the modification should be applied to the target process.
%\parared{Lisp in Lisp}
In addition to the formal definition of computational reflexivity, we 
%describe
provide
 a 
proof-of-concept prototype, implemented through the modification of a meta-circular interpreter. It allows us to 
demonstrate that the proposed mechanism is computationally feasible and even achievable with a small set of instructions.  

%\parared{Inspiration by consciousness studies}

In our definition of computational reflexivity, we have been inspired by several concepts discussed in the literature on consciousness studies. Some of them will be reported in the following sections. Our main source of inspiration is, however, the notion of self-conscious reflexivity, as discussed in higher-order theories of consciousness, and the attempts to describe it in neuroscientific \cite{Damasio1999} and computational \cite{Trautteur2004} terms.

%\parared{Paper outline}

The rest of the paper is organized as follows.
In the next section, we present an overview of self-awareness, introspection, and reflexion in the context of both computer science and consciousness studies.
Section \ref{computational-reflexivity} introduces the formal definitions of computational reflexion, and Section \ref{implementation} introduces the prototype. 
Finally, we present a short discussion in Section \ref{discussion} and draft 
possible applications and next research steps in Section \ref{conclusion}.

%-------------------------------------------
\section{Background}
\label{background}

%\todop{Add here a short introduction to the following subsections.}

%-------------------------------------------
\subsection{Procedural Introspection}
\label{procedural-introspection}

%\parared{Definition of \emph{computational introspection}}

In the context of the present work, we use the term \emph{computational introspection} to indicate a program capable of accessing itself, create a self-representation, and manipulate it.
%\parared{Distinction between ``procedural knowledge'' and ``declarative knowledge''}
A crucial distinction should be made between the meaning of ``knowledge'' underlying the notion of ``representation'' and ``manipulation''.
%More specifically, 
For this reason,
we distinguish between \emph{procedural knowledge} and \emph{declarative knowledge}, the former based on computable functions, and the latter on logical statements.
%\parared{Distinction between ``procedural introspection'' and ``declarative introspection''}
Depending on which meaning of ``knowledge'' is adopted, there are two different 
%definitions of 
ways to define
computational introspection, called here \emph{procedural introspection} and \emph{declarative introspection}, respectively.

%\parared{Definition of ``procedural introspection''}

Batali \cite{Batali1983} claims that ``introspection is the process of thinking about one's own thoughts and feelings. [...] To the degree that \emph{thoughts} and \emph{feelings} are computational entities, computational introspection would require the ability of a process to access and manipulate its own program and its current context'' (See Valdemir and Neto \cite{Valdemir_Neto2007} on \emph{self-modifying code}). 
%This statement corresponds to our definition of procedural introspection, where the \emph{access} corresponds to processing the program as data, and \emph{manipulation} means program modification. 
In other words, computational introspection corresponds to the ability of a program to process its own code as data and modify it\footnote{In this definition, we put together self-representation and self-modification and, thus, the \emph{introspection} and \emph{reflection} features mentioned in Section \ref{introduction}.}.

%\parared{Definition of ``declarative introspection''}

%On the other hand, 
By contrast,
in declarative introspection, the \emph{access} corresponds to the generation of a set of logical statements, while their \emph{manipulation} is performed by logical inference \cite{McCarthy1959}\cite{Weyhrauch1980}. 
 Batali \cite{Batali1983} says that ``The general idea is that a computational system (an agent preferably) embodies a theory of reasoning (or acting, or whatever). This is what traditional Al systems are -- each system embodies a theory of reasoning in virtue of being the implementation of a program written to encode the theory.''
 
%\parared{Self-knowledge and introspection in early days of AI} 
 
As discussed by Cox \cite{Cox2005}, ``From the very early days of AI, researchers have been concerned with the issues of machine self-knowledge and introspective capabilities. Two pioneering researchers, Marvin Minsky and John McCarthy, considered these issues and put them to paper in the mid-to-late 1950’s.  [...] Minsky's \cite{Minsky1968}  contention was that for a machine to adequately answer questions about the world, including questions about itself in the world, it would have to have an executable model of itself. McCarthy \cite{McCarthy1959} asserted that for a machine to adequately behave intelligently it must declaratively represent its knowledge. [...] Roughly Minsky's proposal was procedural in nature while McCarthy's was declarative.''
On the basis of these ideas, Stein and Barnden performed a more recent work to enable a machine to procedurally simulate itself \cite{Stein_Barnden1995}.

Interestingly, Johnson-Laird \cite{JohnsonLaird1983}, inspired by Minsky, proposes a definition of procedural introspection closer to the concept of computable function. He claims that ``Minsky's formulation is equivalent to a Turing machine with an interpreter that consults a complete description of itself (presumably without being able to understand itself), whereas humans consult an imperfect and incomplete mental model that is somehow qualitatively different.'' According to Smith \cite{Smith1982}, ``the program must have access, not only to its program, but to fully articulated descriptions of its state available for inspection and modification.'' [...] Moreover, ``the program must be able to resume its operation with the modified state information, and the continued computation must be appropriately affected by the changes.''

%\parared{Our definition of ``procedural introspection''}

Unlike the use of `procedural' discussed above, actually consisting of a ``declarative'' representation of the ``procedural knowledge'', we employ the term in a more restrictive way. \emph{Procedural introspection} is here limited to program code access and modification, without any logical modeling and inference. In this way, we want to avoid the possible dependence of a particular declarative modeling from the choices of the human designer, instead focusing on aspects connected to program access and modification. 

%-------------------------------------------
\subsection{Introspection in Consciousness Studies}
\label{introspection-consciousness}

%\parared{Intro}

Historically, all the uses of the term `introspection' in computer science have been influenced by the meaning of the same term in philosophy of mind and, later on, neurosciences and cognitive science.
%\parared{Higher-order theories}
In consciousness studies, introspection is often discussed in the context of the so-called \emph{higher-order} (\emph{HO}) theories, based on the assumption that there are different ``levels'' or ``orders'' of mental states. 
Perceptions, emotions, and thoughts are instances of first-order mental states.
Higher-order mental states are mental states about other mental states. 
For example, a thought about thinking something.
Introspection is considered as ``an examination of the content of the first-order states'' \cite{Overgaard_Mogensen2016}.
It is not clear, however,  if introspection itself is a high-order state or it is involved in the occurrence of first-order states. 

%-------------------------------------------
\subsection{Self-Conscious Reflexivity}
\label{conscious-reflexivity}

%\parared{Central role of reflexion}

Introspection
is not generally considered  the main characteristic of
conscious states. 
In contrast, as
claimed by Peters \cite{Peters2013}, ``consciousness is reflexivity'', where \emph{reflexion} is 
the ``awareness that one is perceiving''.
Unlike other defining characteristics, such as intentionality, reflexivity is the only one that is considered unique to consciousness. 
%\parared{Trautteur: key role of reflexion}
Trautteur remarked that Damasio was the first 
scientist
 to describe reflexion in the context of neuroscience \cite{Trautteur2004}.
%\parared{Damasio's statement}
Damasio's definition of reflexion (referred to by the term \emph{core self}) is reported in the following statement:

\begin{itemize}

%\item 
%\textbf{(Damasio):} 
\item~It is the process of an organism caught in the act of ``representing its own changing state as it goes about representing something else'' (\cite[p. 170]{Damasio1999}).

\end{itemize}

This definition 
is meant to be based on biological (and, thus, physicalist, objective) terms since the term `representation' here denotes specific neural patterns.
%\parared{Trautteur's statement}
The next statement 
expresses
the
attempt by Trautteur to translate the above ``metaphorical'' definition in computational terms:

\begin{itemize}

%\item
%\textbf{(Trautteur):} 
\item~[It] is the process of an agent ``processing its own processing while processing an input\footnote{This statement is extracted from unpublished notes by Trautteur.}.''

\end{itemize}

In this version, 
the \emph{organism} is reformulated as a computational \emph{agent} and \emph{representation} 
as a computational  \emph{process}.
%\parared{The identity paradox}
Both the above statements 
present a logical issue. We refer to it as the \emph{identity paradox}. It consists of the fact that the object and the subject of the experience are perceived as the same entity.
It is a \emph{violation of the identity principle}, also detectable in other expressions used by the same and other authors such as ``presence o the self to itself'' or ``the identity of the owner (of experience) and the owned'' \cite{Trautteur2004}.

%-------------------------------------------
\subsection{Elements of Inspiration and Informal Definition of Computational Reflexivity}
\label{inspiration}

%\parared{From identity to simultaneity}

%In order to 
To
overcome this logical contradiction, in the present research we moved the focus from \emph{identity} to \emph{simultaneity}. This frame shifting was inspired by Van Gulick \cite{VanGulick2014}, which emphasizes the simultaneity of observed and observer: ``what makes a mental state M a conscious mental state is the fact that it is accompanied by a simultaneous and non-inferential higher-order (i.e., meta-mental) state whose content is that one is now in M''. 
%Therefore, we may see reflexion 
The above statement triggered the insight that reflexion can be seen
as the simultaneous occurrence of two \emph{distinct} and \emph{synchronized} \emph{processes}.
%There
It implies
three underlying assumptions in it: \emph{temporal extension} (i.e., `state' means that we are dealing with \emph{processes}), \emph{distinction} (i.e., we have \emph{two} separate processes), and \emph{synchronicity} (i.e., the two processes are \emph{simultaneous}).
Because of the temporal extension, the term `simultaneity' is employed here in the sense of \emph{interval simultaneity}, which refers to sequences of events \cite{Jammer2006}.
%However, interval 
Interval
simultaneity does not necessarily imply, 
%in this context, 
here, 
the simultaneity of the single events.
Our assumption of synchronicity requires that each step in one of the two processes must occur only after a corresponding step in the other one. As shown in the next section, each pair of steps are part of the same interpretation loop.

%\parared{Computational reflexion}

Using the second statement as a reference, we informally define \emph{computational reflexion} as the concurrent (i.e. at every step of the interpretation loop) and synchronized execution of a computer program
and manipulation of its code. Correspondingly, an interpreter capable of performing computational reflexion is said to be provided with \emph{computational reflexivity}. This definition implies  that computational reflexivity is a characteristic of a particular class of universal machines.
%\todop{Summarize the statements used as a source of inspiration for the definition of computational reflexivity}
%\parared{Introduction to the next sections}
%In the next two sections, we will provide a formal definition and a specific implementation of computational reflexion. %\todo{consider to remove it}

%-------------------------------------------
\section{Formal Definition of Computational Reflexion}
\label{computational-reflexivity}

%\parared{Intro - Section content}

In this section, we provide, a step a time, all building blocks for the formal definition of computational reflexion. 
%\parared{Modularity as motivation for focusing on the interpreter}
We assume reflexivity as a property applicable to the execution of any computer program, instead of a property of a single program. For this reason, it must rely on a particular type of program interpretation. From the point of view of an interpreter, the execution of a program can be reduced to a number of iterations of the same \emph{interpretation loop}. We use the term \emph{step} to denote a single occurrence of the interpretation loop, despite its internal complexity. We unravel below the definition of computational reflexivity as a sequence of incremental enrichments of the interpretation loop. Each enrichment, referred by both a textual symbol and a graphic mark, is meant to induce a corresponding modification at the process level.

%\parared{Lower step and standard execution}

\vspace{3mm}

\noindent \textbf{1. Lower Step and Standard Execution} \hspace{1mm} The original computational step (i.e., the unmodified interpretation loop) is called here \emph{lower step}, indicated by the symbol $(S_L)$ and the graphic mark \includegraphics[width=0.03\textwidth]{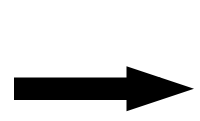}.
At the process level, we call \emph{target process} the overall program execution.

%\parared{Single introspection and tracing}

\vspace{3mm}

\noindent \textbf{2. Single Introspection and Tracing}  \hspace{1mm} In this modified step, the interpreter executes a \emph{local procedural introspection} on the current step, returning the code of the current instruction. It is called \emph{single introspection}, indicated by the symbol $(S_L, I_S)$ and the graphic mark of the interpretation loop is \includegraphics[width=0.03\textwidth]{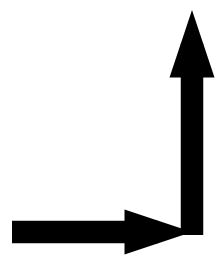}.
At the process level, the system generates a trace of execution, similar to the one produced by a debugger.

%\parared{Single upper step and mirroring}

\vspace{3mm}

\noindent \textbf{3. Single Upper Step and Mirroring}  \hspace{1mm} The interpreter executes the instruction just extracted by introspection. We call it \emph{upper step}, denoted by $(S_L, I_S, S_{SU})$. The overall loop is graphically represented as \includegraphics[width=0.04\textwidth]{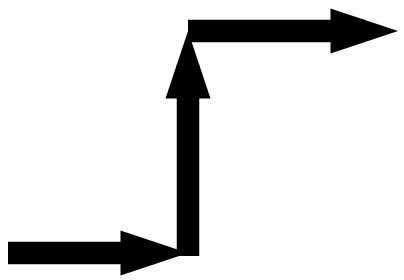}. At the process level, we have two identical programs simultaneously executed. We use the term \emph{mirroring} to indicate this real-time duplication of the target process.

%\parared{Double upper step and augmentation}

\vspace{3mm}

\noindent \textbf{4. Double Upper Step and Augmentation}  \hspace{1mm} Here the interpretation loop is enriched with two further operations: the modification of the current step of the ``mirrored program'' by introduction of an additional instructions, and the next step execution\footnote{Although a more general class of code modification is conceivable, we limit the focus on the modification by instruction insertion. As explained in the next point, the aim is to enrich the second process with information about the target process.}. The term \emph{double upper step}, with the symbol $(S_L, I_S, S_{DU})$, indicates the execution of the ``mirrored'' instruction and the additional one. The overall loop is graphically represented as \includegraphics[width=0.04\textwidth]{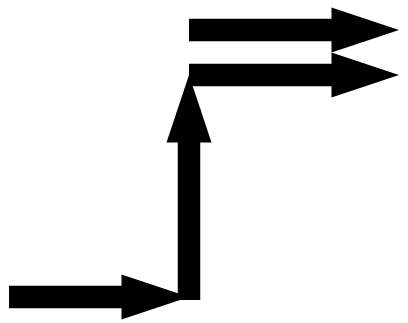}. We call \emph{computational augmentation} the modification of the interpretation loop performed so far. Correspondingly, we have two simultaneous processes: the target process and the augmented process. The latter one is based on the former but modified 
\emph{at the step level}.

%\parared{Double introspection and reflexion}

\vspace{3mm}

\noindent \textbf{5. Double Introspection and Reflexion}  \hspace{1mm} Now, we consider a particular type of computational augmentation, in which the additional instruction of the \emph{double upper step} is a further operation of \emph{global procedural introspection}. While the local introspection returns the code of the current instruction of the target program (i.e., the lower step defined above), the global introspection returns the code of the entire target program or a subset of it. In this case, the upper step consists of an execution of the mirrored instructions of the target program \emph{plus} additional \emph{global} instructions about it. We call \emph{double introspection} this type of double upper step, and denote it by the symbol $(S_L, I_D, S_{DU})$. The overall loop is represented by the graphical mark \includegraphics[width=0.04\textwidth]{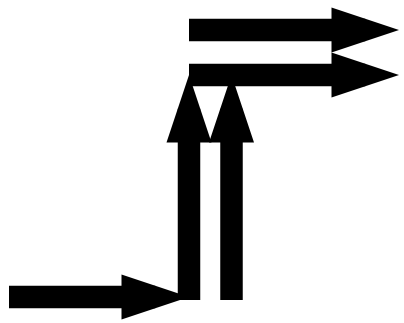}. Finally, we define \emph{computational reflexion} as the process generated by the loop composed by \emph{lower step}, \emph{double introspection} and \emph{double upper step}.

\vspace{3mm}

%\parared{Summary of local and global modifications}

Table \ref{tab:symbols} summarizes the schema of all components. Each row reports the symbolic representation, the graphical mark, and the corresponding terminology at both step and process level. In summary, the addition of specific groups of instructions to the interpretation loop underlies the generation of different processes, each built on the previous one: \emph{standard execution}, \emph{tracing}, \emph{mirroring}, \emph{augmentation}, and \emph{reflexion}. Given a target process, the enriched interpreter executed the related program and a concurrent version executed, at every step, with its own code.
%\parared{Formal definition of computational reflexion}
Our definition of computational reflexion is thus a formal specification 
of the informal one reported in Section \ref{inspiration}.

\begin{table}[ht]
\begin{center}
%\scalebox{0.82}{	
\scalebox{0.98}{	
\renewcommand{\arraystretch}{1.5}
\begin{tabular}{|c|c|c|c|c|}
\hline
\textbf{Symbol} & \textbf{Step Components} & \textbf{Process Creation} & \textbf{Process} \\
\hline\hline
$(S_L)$ \includegraphics[width=0.04\textwidth]{images/execution.png} & \emph{Lower Step} & \emph{Standard Execution} & \emph{Target Process} \\
\hline
$(S_L, I_S)$ \includegraphics[width=0.04\textwidth]{images/proto-self.png} & + \emph{Single Introspection} & \emph{Tracing} & \emph{Execution Trace} \\
\hline
$(S_L, I_S, S_{SU})$ \includegraphics[width=0.06\textwidth]{images/mirroring.png} & + \emph{Single Upper Step} & \emph{Mirroring} & \emph{Mirror Process} \\
\hline
$(S_L, I_S, S_{DU})$ \includegraphics[width=0.06\textwidth]{images/augmentation.png} & $\rightarrow$ \emph{Double Upper Step} & \emph{Augmentation} & \emph{Augmented Process} \\
\hline
$(S_L, I_D, S_{DU})$ \includegraphics[width=0.06\textwidth]{images/reflexion.png} & $\rightarrow$ \emph{Double Introspection} & \emph{Reflexion} & \emph{Reflexive Process} \\
\hline
\end{tabular}
}
\end{center}
\caption{Different versions of the interpretation loop, with the addition of step components, and related computational processes.}
\label{tab:symbols}
\end{table}

%-------------------------------------------
\section{Prototypical Implementation}
\label{implementation}

%\parared{Proof-of-concept implementation}

As a proof of concept of the feasibility to implement computational introspection, as defined in the previous section, we developed a prototypical version. Specifically, we employed and modified the code of a Lisp meta-circular interpreter \cite{Landauer_Bellman2001}\cite{Graham2002} (i.e., an interpreter of the Lisp programming language, implemented in the same language), called here \emph{Lisp in Lisp}.
%\parared{Motivation for using Lisp interpreter}
The main reason for using \emph{Lisp in Lisp} is that it is one of the simplest ways to implement a general-purpose interpreter. 
%In fact, 
Indeed,
it is a specific model of computation based on Church's Lambda Calculus \cite{Church1941, McCarthy1960}. As reported by McCarthy \cite{McCarthy1978}, ``Another way to show that Lisp was neater than Turing machines was to write a universal Lisp function and show that it is briefer and more comprehensible than the description of a universal Turing machine. This was the Lisp function eval [...]'' The program is just a few lines of code and the definition of its main function, \emph{\texttt{eval}}, is based on the composition of a few primitive operators. The \emph{\texttt{eval}} function is what is performing the interpretation (or \emph{evaluation}) process.

%\parared{Definition of interpretation loop}

In this case, we 
%define 
call
\emph{computational step} (and, equivalently, interpretation loop) 
%as 
the \emph{Lisp in Lisp} execution between two next calls of the \emph{\texttt{eval}} function. 
%\parared{Interpreter modification}
Therefore, using the sequence of steps described in the previous section, we modified the definition of \emph{\texttt{eval}} adding additional function calls. For example, the single introspection event correspond to a call of the function \emph{\texttt{quote}}, which returns the code of the argument (i.e. the instruction under execution). 
%\parared{Reference to the web site}
The complete code of the program and applied examples of executions are free available for research purpose%\footnote{We uploaded the material on a web site not provided here for anonymity reasons. We provide it instead as integrative material.}.
% DO NOT REMOVE THE COMMENTED CODE BELOW!
%available at the URL \href{http://valitutti.it/papers/reflexion/index.html}{\textcolor{blue}{\texttt{valitutti.it/papers/reflexion/index.html}}}.

%-------------------------------------------
\section{Discussion}
\label{discussion}

%\parared{Paper contribution - Intro}

The 
%ideas discussed 
intuitions formalized
in this paper are aimed to envision a new type of self-aware systems. While almost all state-of-the-art systems are based on introspection, we propose to consider reflexion as the main aspect of self-awareness. 
%\parared{Intuitive definition of computational reflexion}
We could intuitively define computational reflexion as \emph{``a mechanism for making a computational process continuously self-informed''}. The expression \emph{``mechanism of making''} expresses the fact that reflexion is defined as a particular type of interpreter. Indeed, we focused on the interpretation loop and modified it. Reflexion is not the property of a specific class of computer programs but, instead, something that can be provided to any executable programs through this form of interpretation. Through reflexion, the standard program execution (i.e., the \emph{target process}) is dynamically ``reflected'' into the execution of its augmented counterpart (i.e., the \emph{reflexive process}). 

 As explained in Section \ref{computational-reflexivity}, each instruction of the target program is executed twice: the first time (as sequence of \emph{lower steps}) to achieve the standard execution (and generate the target process), and the second time (as sequence of \emph{upper steps}) as part of the reflexive process. In the above definition, the term \emph{``self''} is not referring to a single entity but to a couple of mutually interactive entities. This \emph{duality} between the two processes is the way we theoretically address  the \emph{identity paradox} mentioned in Section \ref{conscious-reflexivity}.

%------------------------------------------
\section{Possible Applications and Future Work}
\label{conclusion}

%\parared{Possible applications}

The properties identified in the previous section allow us to conceive some interesting uses of the reflexive augmentation of program execution. For example, we could see the execution of the target program and the corresponding reflexive augmentation as performed by two separate but synchronized devices. Specifically, we could have an autonomous agent (e.g. a robot in a physical environment) and an interfaced web service implementing reflexion. Therefore, computational reflexion could be used as a way to provide a system with a temporary ``streaming of self-awareness''.

%\parared{Future work}

The aimed next steps of our research are focused on the following aspects.
Firstly, we intend to further develop the proposed formalization and achieve possible interesting implications as formal theorems.
Secondly, we aim to study the degree to which the reflexive process should give feedback to the target process and modify the related program. In other words, we would like to investigate aspects of run-time ``virtual'' self-modification, not yet taken into account, at this stage of the research, in our prototype. %\todo{virtual self-modification is the crucial property, as claimed in the Introduction; we need to provide an example in the code and modify the above statement}
%According to Damasio, in the human brain, the interaction between first-order and second-order neural representations is different whether the target representation is referred to an internal (i.e. bodily) state or an external perception.
%For example, while the second-order representation of visual perception does not provide feedback to the original process, in the case of internal perception the feedback is needed as part of the emotional response.
%If translated in the ``language of reflexion'' developed here, this statement suggests that, if the target process implements the representation of an agent's internal state or, specifically, an emotional response, it should be modified, to some degree, by the corresponding reflexive process.

A crucial issue is about efficiency. We need to investigate to what degree the combination of step-level local and global introspection and corresponding execution can be feasible performed. If the target program is sufficiently complex, there is a limitation in the 
%amount 
number
of instructions capable of being executed along the duration of the interpretation loop. In this case, the procedural modeling of the target process should be optimized.
%Finally, we intend to study a way to connect a machine-learning to the reflexive module, in such a way to perform an ``offline'' processing of the reflexive process, and use this information to achieve a form of self-organizing behavior.
Finally, we intend to investigate the extent to which computational reflexivity could be employed to achieve a form of self organization, using the information gathered by the step-level introspective acts to train a self-reinforcement system.

%-------------------------------------------
%\section*{References}

%\bibliography{biblio}

%% This BibTeX bibliography file was created using BibDesk.
%% http://bibdesk.sourceforge.net/

%% Created for Alessandro Valitutti at 2017-06-11 19:07:22 +0200 

%% Saved with string encoding Unicode (UTF-8) 

%-------------------------------------------
%\pagebreak
\section*{Appendix A: Code Description}

The following integrative material is provisionally provided as appendix of the paper. In the final version, this material will be provided in a
% website (not yet provided for anonymity reason).
 website.

\noindent\rule[0.5ex]{\linewidth}{1pt}

%\parared{Use of \emph{Lisp in Lisp}}

We employed the code of an interpreter of the Lisp programming language, implemented in the same language, and called here \emph{Lisp in Lisp}. Specifically, since the term \emph{Lisp} is currently used to denote an entire family of programming languages sharing common characteristics, we tested and run the code in Common Lisp \cite{Steele1990}.
Rather than the original formulation of \emph{\texttt{eval}} by McCarthy \cite{McCarthy1960}, we adopted the simpler version by Paul Graham \cite{Graham2002}, which also found a bug in the original version and removed it\footnote{We have found a small bug in Graham's code as well. In the definition body of the \emph{$<$\texttt{pair.}$>$} function, there is a call to the \emph{$<$\texttt{list}$>$} function, which is a system function. Since the set of primitive operators should not include \emph{$<$\texttt{list}$>$}, we defined the function \emph{$<$\texttt{list.}$>$} and 
used it to replace all the occurrences of 
\emph{$<$\texttt{list}$>$}
%replaced it 
in the definition body of \emph{$<$\texttt{pair.}$>$} In this note, we rounded the function names with angular parenthesis to separate them more clearly from the rest of the text.}.

%\parared{Interpretation loop and code of \emph{\texttt{eval}}}

In this context, we define \emph{computational step} (and, equivalently, interpretation loop) as the \emph{Lisp in Lisp} execution between two next calls of the \emph{\texttt{eval}} function.
Appendix \ref{lisp-code} contains
the code of the \emph{Lisp in Lisp} (i.e., the \texttt{eval.} function) and its modified version (called \texttt{eval-augment}) implementing reflexion. 

%-------------------------------------------
\section*{Appendix B: Components of Computational Reflexion}

In the same way proceeded in the previous section, we focused on the interpretation loop and gradually enriched to obtain the version implementing computational reflexion.

%\parared{Steps and processes}

\begin{enumerate}

\item \textbf{Lower Step}. \hspace{1mm} It is equivalent to the interpretation loop defined above. 

\item \textbf{Single (Local) Introspection}. \hspace{1mm} The current \emph{\texttt{eval}} call returns the code of the current instruction, consisting of a function call.

\item \textbf{Single Upper Step}. \hspace{1mm} The code of the current function call, produced by the local introspection, is in turn executed (i.e. \emph{\texttt{eval}} is called on it).

\item \textbf{Double Upper Step}. \hspace{1mm} The code generated by local introspection is enriched with additional instructions. As an example, we added a \emph{\texttt{print}} call to the output of the current call. In this way, the interpreter will display on the terminal the trace of execution.

\item \textbf{Double Introspection}. \hspace{1mm} Finally, the interpretation loop is enriched with the instruction for global introspection. In other words, it returns the code of the entire program.
	
\end{enumerate}

In summary, at any stage of the computation, the interpreter 
%can access 
accesses and executes
the code both locally and globally. In particular, the program code could be modified at each step and, thus, influence the next execution.

As a specific example, Appendix \ref{lisp-examples} shows the Lisp definition of the function \emph{\texttt{my-last}}, which gets a list as input and returns its last element as output.

%-------------------------------------------
%\pagebreak

\section*{Appendix C: Lisp Code} \label{lisp-code}

The code of the function \textbf{\texttt{eval.}} corresponds to the version of the \emph{Lisp in Lisp} by Paul Graham \cite{Graham2002}.
We modified it and defined \textbf{\texttt{eval-augment}}, as a proof-of-concept version of the reflexive interpreter, with the following instruction:

\vspace{-0.3cm}

\begin{center}
	
\textbf{\texttt{(augment input output pred)}}.

\end{center}

The function \textbf{\texttt{augment}} applies the predicate \textbf{\texttt{pred}} to the input and output of the current step. 
The specific implementation of \textbf{\texttt{pred}} in this example is \textbf{\texttt{*pred*}}, which extract the code of the current instruction and execute it again, thus performing the ``mirroring'' discussed in Section 3 of the paper.

\begin{lstlisting}[language=lisp, basicstyle=\footnotesize]

(defun eval. (e a)
  (cond
    ((atom e) (assoc. e a))
    ((atom (car e))
     (cond
       ((eq (car e) 'quote) (cadr e))
       ((eq (car e) 'atom)  (atom   (eval. (cadr e) a)))
       ((eq (car e) 'eq)    (eq     (eval. (cadr e) a)
                                    (eval. (caddr e) a)))
       ((eq (car e) 'car)   (car    (eval. (cadr e) a)))
       ((eq (car e) 'cdr)   (cdr    (eval. (cadr e) a)))
       ((eq (car e) 'cons)  (cons   (eval. (cadr e) a)
                                    (eval. (caddr e) a)))
       ((eq (car e) 'cond)  (evcon. (cdr e) a))
       ('t (eval. (cons (assoc. (car e) a)
                        (cdr e))
                  a))))
    ((eq (caar e) 'label)
     (eval. (cons (caddar e) (cdr e))
            (cons (list. (cadar e) (car e)) a)))
    ((eq (caar e) 'lambda)
     (eval. (caddar e)
            (append. (pair. (cadar e) (evlis. (cdr e) a))
                     a)))))

(defun null. (x)
  (eq x '()))

(defun and. (x y)
  (cond (x (cond (y 't) ('t '())))
        ('t '())))

(defun not. (x)
  (cond (x '())
        ('t 't)))

(defun append. (x y)
  (cond ((null. x) y)
        ('t (cons (car x) 
		  (append. (cdr x) y)))))

(defun list. (x y)
  (cons x (cons y '())))

(defun pair. (x y)
  (cond ((and. (null. x) (null. y)) '())
        ((and. (not. (atom x)) (not. (atom y)))
         (cons (list. (car x) (car y))
               (pair. (cdr x) (cdr y))))))

(defun assoc. (x y)
  (cond 
   ((null. y) '())
   ((eq (caar y) x) (cadar y))
   ('t (assoc. x (cdr y)))))

(defun evcon. (c a)
  (cond ((eval. (caar c) a)
         (eval. (cadar c) a))
        ('t (evcon. (cdr c) a))))

(defun evlis. (m a)
  (cond ((null. m) '())
        ('t (cons (eval.  (car m) a)
                  (evlis. (cdr m) a)))))

(defun eval-augment (e a pred)
  (let*
      ((input (list e a))
       (output
	(cond	
	 ((atom e) (assoc. e a))
	 ((atom (car e))
	  (cond
	   ((eq (car e) 'quote) (cadr e))
	   ((eq (car e) 'atom)  (atom   (eval-augment (cadr e) a pred)))
	   ((eq (car e) 'eq)    (eq     (eval-augment (cadr e) a pred)
					(eval-augment (caddr e) a pred)))
	   ((eq (car e) 'car)   (car    (eval-augment (cadr e) a pred)))
	   ((eq (car e) 'cdr)   (cdr    (eval-augment (cadr e) a pred)))
	   ((eq (car e) 'cons)  (cons   (eval-augment (cadr e) a pred)
					(eval-augment (caddr e) a pred)))
	   ((eq (car e) 'cond)  (evcon. (cdr e) a))
	   ('t (eval-augment (cons (assoc. (car e) a)
			    (cdr e))
		      a pred))))
	 ((eq (caar e) 'label)
	  (eval-augment (cons (caddar e) (cdr e))
		      (cons (list. (cadar e) (car e)) a) pred))
	 ((eq (caar e) 'lambda)
	  (eval-augment (caddar e)
		      (append. (pair. (cadar e) (evlis. (cdr e) a)) 
			       a) pred)))))
    (augment input output pred)
    output))


(defun augment (input output pred)
  (setq *done* (append *done* (list (list input output))))
  (funcall pred *done*))


(setq *pred* #'(lambda (done)
		  (let*
		      ((next (car (last done)))
		       (input (car next))
		       (e (car input))
		       (a (cadr input))
		       (output1 (eval. e a)))
		    (format t 
			    (concatenate 'string
				      (write-to-string input) "~%"
				      "-> " (write-to-string output1) "~%"))
		 t)))

\end{lstlisting}

%-------------------------------------------
\pagebreak

\section{Appendix E: Examples of Execution} \label{lisp-examples}

As a simple example, the function \textbf{\texttt{augment}} and the predicate \textbf{\texttt{*pred*}} are applied to simple data (the atom \textbf{\texttt{a}} with value \textbf{\texttt{1}}, and the function \textbf{\texttt{car}} returning the first element of the list \textbf{\texttt{(a b)}}. In particular, it is applied to the simple recursive function \textbf{\texttt{my-last}}, returning the last element of a list.
 
\begin{lstlisting}[language=lisp, basicstyle=\footnotesize]

CL-USER(356): (eval-augment 'a '((a 1)) *pred2*)
1
CL-USER(357): (eval-augment 'a '((a 1)) *pred1*)
(A ((A 1)))
-> 1
1


CL-USER(360): (eval. '(car '(a b)) nil)
A
CL-USER(361): (eval-augment '(car '(a b)) nil *pred2*)
A
CL-USER(362): (eval-augment '(car '(a b)) nil *pred1*)
('(A B) NIL)
-> (A B)
((CAR '(A B)) NIL)
-> A
A


CL-USER(370): (setq e '(my-last '(a b c)))
(MY-LAST '(A B C))
CL-USER(371): (setq a '(
     (my-last (label my-last 
		     (lambda (x) 
		       (cond 
			((null. x) 'nil) 
			((null. (cdr x)) (car x)) 
			('t (my-last (cdr x)))
			)))) 
     (null. (label null. (lambda (x) (eq x nil))))
     ))
((MY-LAST (LABEL MY-LAST (LAMBDA (X) (COND # # #))))
 (NULL. (LABEL NULL. (LAMBDA (X) (EQ X NIL)))))
CL-USER(372): (eval. e a)
C
CL-USER(373): (eval-augment e a *pred2*)
C
CL-USER(374): (eval-augment e a *pred1*)
((COND ((NULL. X) 'NIL)
       ((NULL. (CDR X)) (CAR X))
       ('T (MY-LAST (CDR X))))
 ((X (A B C))
  (MY-LAST
   (LABEL MY-LAST
    (LAMBDA (X)
      (COND ((NULL. X) 'NIL)
            ((NULL. (CDR X)) (CAR X))
            ('T (MY-LAST (CDR X)))))))
  (MY-LAST
   (LABEL MY-LAST
    (LAMBDA (X)
      (COND ((NULL. X) 'NIL)
            ((NULL. (CDR X)) (CAR X))
            ('T (MY-LAST (CDR X)))))))
  (NULL. (LABEL NULL. (LAMBDA (X) (EQ X NIL))))))
-> C
(((LAMBDA (X)
    (COND ((NULL. X) 'NIL)
          ((NULL. (CDR X)) (CAR X))
          ('T (MY-LAST (CDR X)))))
  '(A B C))
 ((MY-LAST
   (LABEL MY-LAST
    (LAMBDA (X)
      (COND ((NULL. X) 'NIL)
            ((NULL. (CDR X)) (CAR X))
            ('T (MY-LAST (CDR X)))))))
  (MY-LAST
   (LABEL MY-LAST
    (LAMBDA (X)
      (COND ((NULL. X) 'NIL)
            ((NULL. (CDR X)) (CAR X))
            ('T (MY-LAST (CDR X)))))))
  (NULL. (LABEL NULL. (LAMBDA (X) (EQ X NIL))))))
-> C
(((LABEL MY-LAST
   (LAMBDA (X)
     (COND ((NULL. X) 'NIL)
           ((NULL. (CDR X)) (CAR X))
           ('T (MY-LAST (CDR X))))))
  '(A B C))
 ((MY-LAST
   (LABEL MY-LAST
    (LAMBDA (X)
      (COND ((NULL. X) 'NIL)
            ((NULL. (CDR X)) (CAR X))
            ('T (MY-LAST (CDR X)))))))
  (NULL. (LABEL NULL. (LAMBDA (X) (EQ X NIL))))))
-> C
((MY-LAST '(A B C))
 ((MY-LAST
   (LABEL MY-LAST
    (LAMBDA (X)
      (COND ((NULL. X) 'NIL)
            ((NULL. (CDR X)) (CAR X))
            ('T (MY-LAST (CDR X)))))))
  (NULL. (LABEL NULL. (LAMBDA (X) (EQ X NIL))))))
-> C
C

\end{lstlisting}

\end{document}